\documentclass[letterpaper]{article}
\usepackage{aaai}
\nocopyright 
\usepackage{subfig}
\usepackage{times}
\usepackage{helvet}
\usepackage{courier}
\usepackage[numbers]{natbib}
\usepackage{amsthm}
\usepackage{amssymb,amsfonts}
\usepackage[fleqn]{amsmath}

\usepackage[export]{adjustbox}
\usepackage{algorithm,algorithmic}
\usepackage{hyperref}
\usepackage{enumitem}
\usepackage{graphicx}
\usepackage{textcomp}
\usepackage{bbm}
\usepackage[utf8]{inputenc}
\usepackage{xcolor}
\usepackage{comment}
\newtheorem{definition}{Definition}

\DeclareMathOperator*{\argmin}{arg\,min}
\DeclareMathOperator*{\argmax}{arg\,max}
\newtheorem{theorem}{Theorem}
 \usepackage [autostyle, english = american]{csquotes}
\MakeOuterQuote{"}
\usepackage{amsmath}
\usepackage{adjustbox}
\usepackage{array}
\usepackage{booktabs}
\usepackage{multirow}
\usepackage{hhline}
\begin{document}
%
\title{Fairness-Aware Neural R\'eyni  Minimization for Continuous Features}
\author{Vincent Grari\textsuperscript{1},
Boris Ruf\textsuperscript{2},
Sylvain Lamprier\textsuperscript{1},
Marcin Detyniecki\textsuperscript{2}\textsuperscript{3}\\
\textsuperscript{1}{Sorbonne Universit\'{e}, LIP6/CNRS, Paris, France}\\
\textsuperscript{2}{AXA, REV Research, Paris, France}\\
\textsuperscript{3}{Polish Academy of Science, IBS PAN, Warsaw, oland}\\
vincent.grari@lip6.fr,
boris.ruf@axa.com, 
sylvain.lamprier@lip6.fr,
marcin.detyniecki@axa.com
}

\maketitle
\begin{abstract}
The past few years have seen a dramatic rise of academic and societal interest in fair machine learning. While plenty of fair algorithms have been proposed recently to tackle this challenge for discrete variables, only a few ideas exist for continuous ones. The objective in this paper is to ensure some independence level between the outputs of regression models and any given continuous sensitive variables. For this purpose, we use the Hirschfeld-Gebelein-R\'enyi (HGR) maximal correlation coefficient as a fairness metric. We propose two approaches to minimize the HGR coefficient. First, by reducing an upper bound of the HGR with a neural network estimation of the $\chi^{2}$ divergence. 
Second, by minimizing the HGR directly with an adversarial neural network architecture. The idea is to predict the output $Y$ while minimizing the ability of an adversarial neural network to find the estimated transformations which are required to predict the HGR coefficient. We empirically assess and compare our approaches and demonstrate significant improvements on previously presented work in the field. 

\end{abstract}

\section{Introduction}
The use of machine learning algorithms in our day-to-day life has become ubiquitous. However, when trained on biased data, those algorithms are prone to learn, perpetuate or even reinforce these biases~\citep{Bolukbasi2016}. Because many applications have far-reaching consequences (credit rating, insurance pricing, recidivism score, etc.), there is an increasing concern in society that the use of machine learning models could reproduce discrimination based on sensitive attributes such as gender, race, age, weight, or other. In fact, many incidents of this kind have been reported in recent years. For example, an analysis software producing criminal risk scores in the United States (COMPAS) systematically discriminated against black defendants~\citep{angwin2016machine}. Also, discrimination based on gender can be seen in targeted and automated online advertising for job opportunities in the Science, Technology, Engineering and Math (STEM) fields~\citep{Lambrecht}.

A widely applied method to achieve fairness is to simply remove any sensitive attributes from the data set~\citep{Pedreshi2008}. However, this concept, known as "fairness through unawareness", is highly insufficient because any other non-sensitive attribute might indirectly contain significant sensitive information. For example, the height of an adult could provide a strong indication about the gender.

A new research field has emerged to find solutions to this problem: fair machine learning. Its overall objective is to ensure that the prediction model is not dependent on a sensitive attribute~\cite{Zafar2017mechanisms}. Most of the previously presented work focuses on discrete values. This may not hold when, for instance, the sensitive attribute is age or the output is income. Recently, one paper has discussed fair machine learning for continuous variables, approximating an upper bound of the Hirschfeld-Gebelein-R\'enyi (HGR) correlation coefficient exploiting the Witsenhausen's characterization~\cite{mary2019fairness}. Inspired by this idea, we enhance and improve the approach. Further, we present an adversarial neural network architecture which minimizes the HGR coefficient directly.

The contributions of this paper are:
\begin{itemize}
    \item We present a more generalizable approach to minimize the HGR coefficient by reducing an approximation of the $\chi^{2}$ divergence inspired by Mutual Information Neural Estimation (MINE)~\citep{belghazi2018mutual};
    \item We propose a neural network architecture which minimizes the HGR coefficient with an adversarial approach. The adversarial directly approximates HGR by finding non-linear transformations of the data;
    \item We demonstrate empirically that our neural HGR-based approaches are very competitive for fairness learning with continuous features   
    on artificial and real-world popular data sets.
\end{itemize}

The remainder of this paper is as follows. First, Section~\ref{sec:related_work} reviews papers related with our work. Section~\ref{sec:Fairness_Criterion} introduces different definitions of fairness and metrics. 
Section~\ref{sec:HGR_Adversarial_Network} outlines the architecture of our fair adversarial HGR algorithm, with the two continuous fairness loss we introduce. Finally, Section~\ref{sec:results} discusses the experimental results of our approach.

\section{Related Work}
\label{sec:related_work}
Significant work has been done in the field of fair machine learning recently, in particular when it comes to quantifying and mitigating undesired bias. For the mitigation approaches, three distinct strategy groups exist.

Algorithms of the "pre-processing" group mitigate bias which exists in the training data. The ideas range from suppressing the sensitive attributes, learning fair representations, or changing class labels of the data set to reweighing or resampling the data~\citep{Kamiran2012,zemel2013learning,Bellamy2018,Calmon2017}.

The second group of mitigation strategies comprises the "in-processing" algorithms. For this type of algorithms, unwanted bias gets mitigated during the learning phase. One approach to achieve this objective is to include a fairness constraint directly in the loss function. For example, a decision boundary covariance constraint could be added to logistic regression or linear SVM algorithms~\citep{Zafar2017mechanisms}. A different paper proposes a meta algorithm which expects a fairness metric as part of the input in order to produce a new classifier optimized towards that metric~\citep{DBLP:journals/corr/abs-1806-06055}. Yet another concept is adversarial debiasing, an architecture inspired by generative adversarial networks (GANs)~\citep{NIPS2014_5423}. More precisely, in this approach a traditional classifier is trained to predict the label $Y$, while an adversarial neural network is trained at the same time with the objective to predict a sensitive attribute $S$~\citep{Zhang2018,Wadsworth2018,Louppe2016}.

The final group of mitigation strategies are the "post-processing" algorithms. In this approach, only the output labels of a trained classifier are adjusted. For example, by optimizing for an equalized odds objective, a Bayes predictor model can modify output labels~\citep{Hardt2016}. Another paper proposes a weighted estimator for demographic disparity which makes use of soft classification based on proxy model outputs~\citep{Chen2019}. On the one hand, post-processing algorithms have the advantage that fair classification is achieved without the need to modify or retrain the original model. On the other hand, this concept may negatively impact the accuracy or could affect any generalization retrieved from the original classifier~\citep{Donini2017}.

Most of the present work in fair machine learning focuses on categorical variables with a supervised classification problem and a binary sensitive feature. Recently, an approach for continuous variables using the Witsenhausen's characterization of the R\'enyi correlation coefficient was presented~\cite{mary2019fairness}. They extend the work proposed by \cite{kamishima2011fairness} with the minimization of an estimation of the Mutual Information (MI) for categorical variables. This algorithm is a "in-processing" fairness approach. They minimize the Hirschfeld-Gebelein-R\'enyi (HGR) correlation coefficient by optimizing the $\chi^{2}$ divergence. 
However, they make a strong assumption by basing their approach on a Gaussian Kernel Distribution Estimator (KDE).  This makes it difficult to generalize on all different kinds of data sets. We propose to extend this idea and make it as generalizable as possible by modifying the KDE method by neural estimation and then comparing it with our adversarial algorithm which is detailed in Section~\ref{sec:HGR_Adversarial_Network}.





\section{Fairness Definitions and Metrics}
\label{sec:Fairness_Criterion}

\subsection{Fairness Objectives}

Throughout this document, we consider a supervised machine learning algorithm for regression problems. The training data consists of $n$ examples 
${(x_{i},s_{i},y_{i})}_{i=1}^{n}$, where $x_{i} \in \mathbb{R}^{p}$ is the feature vector with $p$ predictors of the $i$-th example, $s_i$ is its continuous sensitive attribute and $y_{i}$ its continuous outcome. 
We describe below two common fairness definitions 
that we use in this work in the continuous setting.  

\textit{Demographic Parity:} One of the main objectives in fair machine learning is to ensure that the sensitive attribute $S$ is independent of the output predictions $\widehat{Y}$: $\widehat{Y} \perp S$.
Compared to the most common discrete binary setting, where the demographic parity can be reduced to ensure that $E[\widehat{Y}|S]=E[\widehat{Y}]$ \cite{pmlr-v80-agarwal18a}, the continuous case is more complicated since it comes down to consider distribution divergences rather than simple conditional expectations.   


\textit{Equalized Residuals:} A model is considered fair when the residuals $\widehat{Y}-Y$ are independent from the sensitive attribute $S$: $(Y-\widehat{Y}) \perp S$. 
To illustrate it, let's imagine a car insurance pricing scenario where young people have higher claims than older people. A classical pricing model would charge young people a higher premium. 
In the case of demographic parity, the average price must be the same across all ages. This means that older people would generally pay more than their real cost, and younger people less. In contrast, the equalized residuals setting only 
ensures that for all ages, the overall error does not deviate too much.

\subsection{Metrics for the Continuous Setting}

In order to assess these fairness definitions in the continuous case, it is essential to look at the concepts and measures of statistical dependence. There are many methods to measure the dependence between two variables. A simple way is to measure the Pearson's rho, Kendall's tau or Spearman's rank. Those types of measure have already been used in fairness, mitigating the conditional covariance for categorical variables~\cite{Zafar2017mechanisms}. However, the major problem with these measures is that they only capture a limited class of association patterns, like linear or monotonically increasing functions. The Pearson correlation, for instance, only measures the linear relationship. When choosing a single non-linear transformation such as the square function of a uniform distribution between -1 and 1, for example, this coefficient results in a theoretical correlation of 0. 

Over the last few years, many non-linear dependence measures have been introduced like the Kernel Canonical Correlation Analysis (KCCA)~\citep{hardoon2009convergence}, the Distance or Brownian Correlation (dCor)~\citep{szekely2009brownian},  the Hilbert-Schmidt Independence Criterion (HSIC and CHSIC)~\citep{Gretton:2005:KMM:1046920.1194914,poczos2012copula} or the Hirschfeld-Gebelein-R\'enyi (HGR)~\citep{renyi1959measures}. 
Comparing those non-linear dependence measures~\citep{lopez2013randomized}, the HGR coefficient seems to be an interesting choice: It is a normalized measure which is capable of correctly measuring linear and non-linear relationships, it can handle multi-dimensional random variables and it is invariant with respect to changes in marginal distributions.



 
\begin{definition}
For two jointly distributed random variables $U \in \mathcal{U}$ and $V \in \mathcal{V}$
, the Hirschfeld-Gebelein-R\'enyi maximal correlation is
defined as:
\begin{eqnarray}
HGR(U, V) = \sup_{\substack{ f:\mathcal{U}\rightarrow \mathbb{R},g:\mathcal{V}\rightarrow \mathbb{R}\\
           E(f(U))=E(g(V))=0 \\
           E(f^2(U))=E(g^2(V))=1}} \rho(f(U), g(V)) \\
          = \sup_{\substack{ f:\mathcal{U}\rightarrow \mathbb{R},g:\mathcal{V}\rightarrow \mathbb{R}\\
           E(f(U))=E(g(V))=0 \\
           E(f^2(U))=E(g^2(V))=1}} E(f(U)g(V))
\label{hgr}
\end{eqnarray}

where $\rho$ is the Pearson linear correlation coefficient~\footnotemark[1] with some measurable functions $f$ and $g$. 
\end{definition}
The HGR coefficient is equal to 0 if the two random variables are independent. If they are strictly dependent the value is 1.
\footnotetext[1]{$\rho(U, V)$ := $\frac{Cov(U;V)}{\sigma_{U}\sigma_{V}}$, where $Cov(U;V)$, $\sigma_{U}$ and $\sigma_{V}$ are the covariance between $U$ and $V$, the standard deviation of $U$ and the standard deviation of $V$, respectively.}
The dimensional spaces for the functions $f$ and $g$ are infinite. This property is the reason why the HGR coefficient proved difficult to compute. 

In information theory literature, the Witsenhausen's characterization~\cite{witsenhausen1975sequences} proposes a simple approximation of the HGR coefficient for discrete features. It demonstrates the possibility to estimate this coefficient directly by the calculation of the second largest value of a specific matrix ($Q$ below).
It is briefly described in the following:



\begin{theorem}
Let $U$ and $V$ be discrete variables and the matrix $Q$ be defined as follows: 

\begin{eqnarray}
Q_{U,V}(j,j^{'}) = \frac{P_{U,V}(j,j^{'})}{\sqrt{P_{U}(j)}\sqrt{P_{V}(j^{'})}}
\end{eqnarray}
Where $P_{U,V}$ is the joint distribution of $U$ and $V$, $P_{U}$ and $P_{V}$ are the corresponding marginal distributions. \\
Under mild conditions~\cite{witsenhausen1975sequences}: 
\begin{eqnarray}
HGR(U, V) = \sigma_{2}(Q_{U,V})
\end{eqnarray}
where $\sigma_{2}$ is the second largest singular value of a matrix. 
\label{theorem_witsenhausen}

\end{theorem}

It was shown that such a calculation of the HGR coefficient can be used as fairness constraint for discrete variables~\cite{baharlouei2019r}. For continuous variables, however, this proved difficult. An approximation 
can be done with strong assumptions such that the matrix $Q$ is viewed as the kernel of a linear operator on $L^2(dP_{U}dP_{V})$~\cite{witsenhausen1975sequences}. This approximation has been used with Kernel density estimation (KDE) as a fairness metric by \cite{mary2019fairness}. We will refer to this metric in our experiments 
as HGR\_KDE.
Another way to approximate this coefficient is to require that $f$ and $g$ belong to Reproducing Hilbert Kernel's spaces (RKHS) by taking the largest canonical correlation between two sets of copula random projections. This has been done efficiently under the name of Randomized Dependency Coefficient (RDC) \cite{lopez2013randomized}. We will make use of this approximated metric 
as HGR\_RDC. 

\section{Neural HGR Minimization for Fairness} 
\label{sec:HGR_Adversarial_Network}
 As explained in Section~\ref{sec:Fairness_Criterion}, the 
 HGR coefficient can be leveraged for fairness learning. However, it's direct use for training fair models is difficult, especially for the continuous case, since it requires the optimization of the second largest singular value of an estimated 
 matrix ${\cal Q}$ (in the case of HGR\_KDE), or the computation of random non-linear projections and the estimation of copula transformations (in the case of HGR\_RDC), at each step of the learning process. 
 
In this paper, we propose  two novel neural HGR-based costs for fairness in the continuous setting, that can be plugged in the following generic optimization problem: 
\begin{eqnarray}
\begin{aligned}
    \argmin_{{\phi}} & \ \mathcal{L}(h_\phi(X),Y)+
    \lambda \Psi(U,V)
    \label{genericfunction}
\end{aligned}
\end{eqnarray}
where  $\mathcal{L}$ is the regression loss function (the mean squared error in our experiments) between the output $h_\phi(X) \in \mathbb{R}$ and the corresponding target $Y$, with $h_\phi$ a neural network with parameters $\phi$, and $\Psi(U,V)$ is a correlation loss between two variables defined as: $$
\left\{
    \begin{array}{ll}
       U=h_\phi(X) \text{ and } V=S   \mbox{ for demographic parity}; \\
        U=h_\phi(X)-Y \text{ and } V=S   \mbox{ for equalized residuals}.
    \end{array}
\right.
$$
The aim is thus to find a mapping $h_\phi(X)$ which both minimizes the  deviation with the expected target $Y$ and does not imply too much dependency of $U$ with the sensitive $S$, according to its definition for the desired fairness objective. The hyperparameter $\lambda$ controls the impact of the correlation loss in the optimization. The correlation loss $\Psi$ can correspond to a Pearson coefficient or a Mutual Information Neural Estimation (MINE \cite{belghazi2018mutual}), 
or one of the two HGR neural estimators proposed in the following of this section. In any case, the objective function is optimized via stochastic gradient descent.

\subsection{Fairness via Neural $\chi^2$-Divergence}
\label{sec:hgr_minimization}


Since it is hard to compute the HGR coefficient in the continuous case, the approach presented in \citep{mary2019fairness} demonstrates how to relax the HGR minimization with a $\chi^2$ divergence between joint probability distribution and product of the marginal distribution. 



Let $U$ and $V$ be continuous variables:
\begin{eqnarray}
HGR(U,V)^2 \leq \chi^{2}(P_{U,V}, P_{U} \otimes P_{V})
\end{eqnarray}
with 
\begin{eqnarray}
\chi^{2}(P_{U,V}, P_{U} \otimes P_{V}) =  \int_{\mathbb R}\int_{\mathbb R} Q_{U,V}(j,j^{'})^2  djdj' -1
\end{eqnarray}

It would therefore be possible to penalize the learning objective in order to minimize the HGR without directly calculating it, by considering $\Psi(U,V)=\chi^{2}(P_{U,V}, P_{U} \otimes P_{V})$.  
The difficulty then relies in the estimation of the $\chi^2$ divergence, which requires the joint and marginal distributions of the two random variables $U$ and $V$. 

Many recent works focused on the approximation of the mutual information, which is defined as:
\begin{eqnarray}
I(U,V)=  \int_{\mathbb R}\int_{\mathbb R} P_{U,V}(j,j^{'})*log(Q_{U,V}(j,j^{'})) djdj'
\end{eqnarray}
Its estimation can be done with a k-NN-based non-parametric estimator \citep{kraskov2004estimating} or by neural estimation with MINE~\cite{belghazi2018mutual}. 


However, only few works focused on the $\chi^2$ divergence recently. The strategy proposed by ~\citep{mary2019fairness} is to estimate the $\chi^2$ divergence via Kernel Density Estimation (KDE). 
This implies the difficult 
choice of the kernel density function to be used.  All estimations in \citep{mary2019fairness} have been done with a Gaussian kernel by setting the bandwidth with the classic Silverman's rule~\citep{silverman1988}, which can be hard to generalize 
for all data. 

To avoid approximating the density with KDE, we apply an approach similar to 
MINE~\citep{belghazi2018mutual}.
For this, we use the following "dual representation" of $\chi^2$. 

\begin{theorem}
The $\chi^2$ divergence admits the following representation\citep{broniatowski2011estimation}: 
\begin{eqnarray}
\chi^2(P,Q) = \sup_{f: \Omega \rightarrow \mathbb{R}} E_{P}(f) - E_{Q}(f + \frac{1}{4} f^2) 
\label{x2}
\end{eqnarray}
where the supremum is taken over all functions $f$ such that
the expectations are finite.
\end{theorem}



\begin{algorithm}
\caption{$\chi^2$ Neural Estimation}
\begin{algorithmic} 

\STATE \textbf{Input:} Distributions $P_{U,V}$ and $P_V$, Neural Network $f_{\theta}$, {\color{white} \textbf{Input:}} Batchsize $b$, Learning rate $\alpha$
\STATE \textbf{repeat}
\STATE Draw $b$ samples from the joint distribution:
\STATE $(u_{1}, v_{1}), . . . ,(u_{b}, v_{b}) \sim P_{UV}$
\STATE Draw $b$ samples from the $V$ marginal distribution:
\STATE $\bar{v}_{1},..., \bar{v_{b}} \sim P_{V}$
\STATE Evaluate the lower bound:
\STATE $J(\theta) \leftarrow \frac{1}{b} \sum_{i=1}^{b} f_{\theta}(u_{i},v_{i})-\frac{1}{b}\sum_{i=1}^{b}(f_{\theta}(u_{i},\bar{v}_{i})+\frac{1}{4}f_{\theta}(u_{i},\bar{v}_{i})^2)$
\STATE Update the network parameters by gradient ascent:
\STATE $\theta \leftarrow \theta +  \alpha \nabla J(\theta)$
\STATE $\textbf{until}$ convergence
\end{algorithmic}
\label{chi2est}
\end{algorithm}


Like MINE~\cite{belghazi2018mutual} we use a set of functions $\mathcal{F}_{\Theta}=\{f_{\theta}\}_{\theta \in \Theta}$, defined by a given neural network  $f$ with parameters $\theta \in \Theta$, as the class of functions that can be considered in \ref{x2} for the approximation $\chi^2_{\theta}$ of $\chi^2$.  
We get the following objective to optimize for neural estimation of $\chi^2$: 
\begin{multline}
\chi^2_{\Theta}(P_{UV},P_{U} \otimes P_{V}) = \\ \max_{\theta \in \Theta} E_{P_{UV}}(f_{\theta})  -E_{P_{U} \otimes P_{V}}(f_{\theta}+\frac{1}{4}f_{\theta}^2)
\label{chi_square_bound}
\end{multline}

This is done by algorithm \ref{chi2est}, which takes distributions $P_{UV}$ and $P_V$ as input, from which it draws mini-batches of $P_{U,V}$ and $P_{U} \otimes P_{V}$ (in practice, samples are drawn from training data $(U,V)$ rather than from true distributions). Then, mini-batches are used to estimate and optimize the difference of expectations from \ref{chi_square_bound} via stochastic gradient ascent.     

\begin{figure}[h]
  \centering
  \includegraphics[scale=0.45,valign=t]{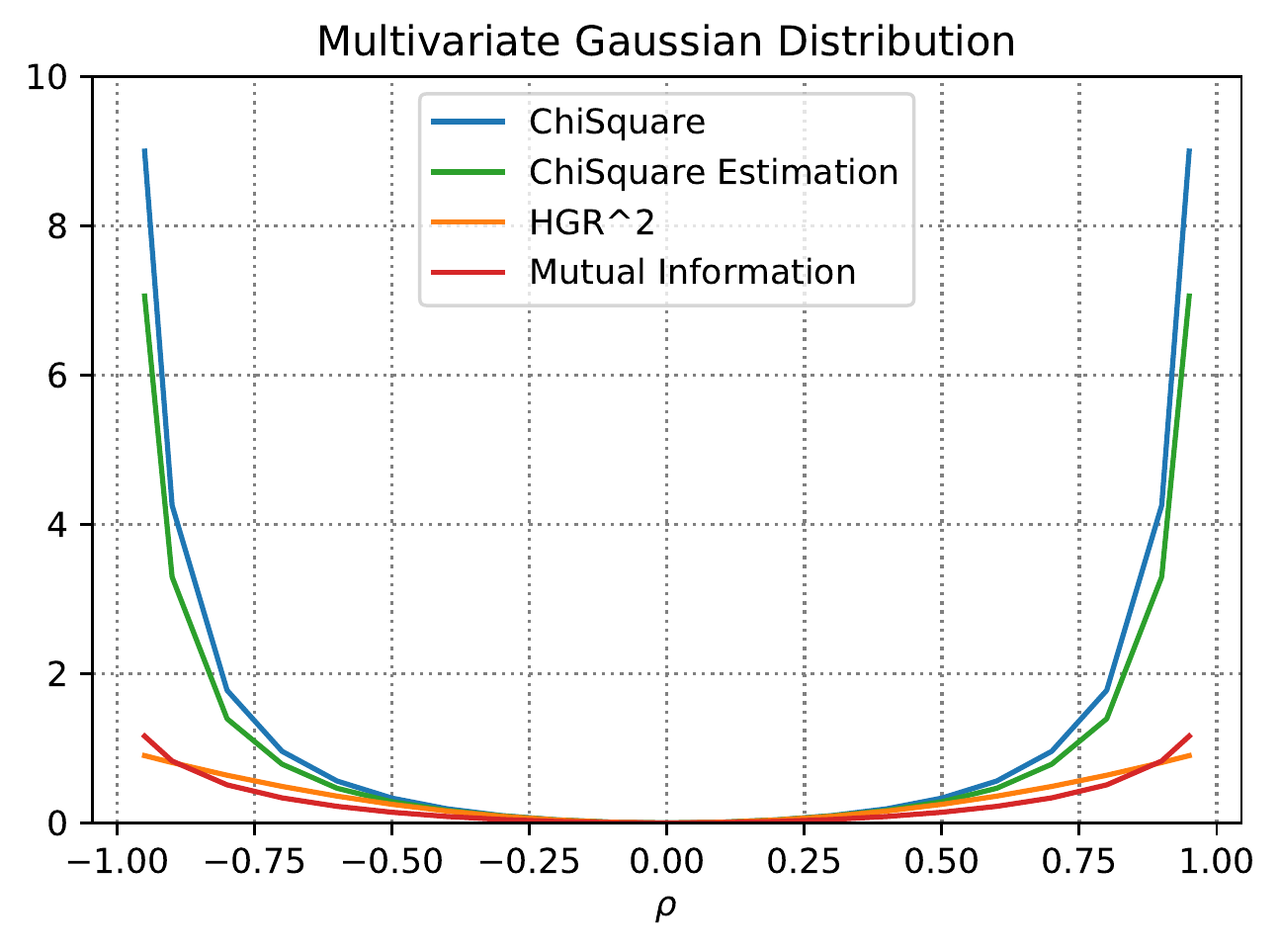}
  \caption{$\chi^2$ estimation for bivariate Gaussians with a $\rho$ correlation coefficient}
  \label{fig:multivariate_gaussians}
\end{figure}
Note that, since the use of neural networks for $\chi^2$ estimation restricts the possible functions to a given compact set $\mathcal{F}_{\Theta}$ defined by the neural architecture used, $\chi^2_{\Theta}$ is only a lower-bound of $\chi^2$: $\chi^2(P_{UV},P_{U} \otimes P_{V}) \geq \chi^2_{\Theta}(P_{UV},P_{U}\otimes P_{V})$. However, it enables to obtain very efficient estimations of $\chi^2$ for various distributions of variables, as shown in figure \ref{fig:multivariate_gaussians} for Multivariate Gaussians where the x-axis represents different covariances between $U$ and $V$. In this figure, we observe that, as expected, the estimated $\chi^2_{\theta}$ divergence is below the real $\chi^2$ for this data, while presenting a very similar shape.  
We also observe that $\chi^2_{\theta}$ always remains above the $HGR^2$, which is the quantity of interest in this paper. For comparison purposes, figure \ref{fig:multivariate_gaussians} also plots the mutual information on this data, for which no such dominance relationship can be established (which can be theoretically verified, see in the supplementary material).

This means that minimizing $\chi^2_{\theta}$ by plugging it as $\Psi(U,V)=\chi^2_{\Theta}(P_{UV},P_{U}\otimes P_{V})$ in our generic objective \ref{genericfunction}, would tend to lower  
the $HGR(U,V)$ correlation 
between $U$ and $V$ for fairness purposes.    
This defines a two-players adversarial min-max game, such as in GANs, where the goal is to find some regression parameters $\phi$ which both minimize the regression loss and the correlation loss, whose computation requires the maximization of the neural parameters $\theta$. All parameters are learned conjointly. As classically done in adversarial learning, we alternate steps for the adversarial maximization (the $\chi^2$ estimation, one iteration of the algorithm \ref{chi2est} at each step) and steps of global loss minimization (one gradient descent iteration on the same batch of data). The detailed Fair $\chi^2$ NN algorithm can be found in the supplementary material.

\newtheorem{abc}{Given any Given any values , of the desired accuracy
and confidence parameters, we have,}

\subsection{Fairness via Neural HGR }

As shown in previous section, reducing the $\chi^2$ divergence leads to reduce the $HGR^2$. However, optimizing this proxy, which corresponds to an upper-bound that is not always very tight, may lead to over-penalize some interesting regression models. 
In the case of bivariate gaussians, it can be observed in Figure \ref{fig:multivariate_gaussians} that the $\chi^2$ curve is widely above the $HGR^2$ one on the extrema, with a dominance that exponentially increases with the absolute covariance of variables $U$ and $V$. 

Another possible approach is to estimate the HGR coefficient directly with neural networks. For this, we use two inter-connected neural networks to approximate the optimal transformations functions $f$ and $g$ from \ref{hgr}. Let $f_{\theta_{f}}$ and $g_{\theta_{g}}$ be two neural networks with respective parameters $\theta_{f}$ and $\theta_{g}$. 
The estimation of HGR can be written as 
the following 
maximization problem: 
\begin{equation}
    HGR_{\Theta}(U,V) = \max_{\theta_f,\theta_g \in \Theta}  E[\hat{f}_{\theta_{f}}(U)\hat{g}_{\theta_{g}}(V)] 
\label{opthgr}
\end{equation}
with $\hat{f}_{\theta_{f}}$ and $\hat{g}_{\theta_{g}}$ the respective  standardization of outputs  from $f_{\theta_{f}}$ and $g_{\theta_{g}}$ according  to $P_U$ and $P_V$:
$$ \hat{f}_{\theta_{f}}(U)=\frac{f_{\theta_{f}}(U) - m_f}{\sigma_f} \qquad \hat{g}_{\theta_{g}}(V)=\frac{g_{\theta_{g}}(V) - m_g}{\sigma_g}  $$
where $m_f$ (resp. $m_g$) is the expectation $E_U[f_{\theta_{f}}(U)]$ (resp. $E_V[g_{\theta_{g}}(V)]$) and $\sigma^2_f$ (resp. $\sigma^2_g$) is the variance $E_U[f_{\theta_{f}}(U)^2]-E_U[f_{\theta_{f}}(U)]^2$ (resp. $E_V[g_{\theta_{g}}(V)^2]-E_V[g_{\theta_{g}}(V)]^2$) of $f$ (resp. $g$) w.r.t. $P_U$ (resp. $P_V$). The standardization of outputs from $f_{\theta_{f}}$ and $g_{\theta_{g}}$ allows us to ensure the constraints on $f$ and $g$ in (\ref{hgr}).




\begin{algorithm}[h]
\caption{HGR Estimation by Neural Network}
\begin{algorithmic} 
\STATE \textbf{Input: } Distributions $P_{U,V}$, Neural Networks $f_{\theta_f}$ and $g_{\theta_g}$, 
{\color{white} \textbf{Input: }} Batchsize $b$, Learning rates $\alpha_f$, $\alpha_g$  
\STATE \textbf{repeat}
\STATE Draw $b$ samples from the joint distribution: 
\STATE $(u_{1}, v_{1}), . . . ,(u_{b}, v_{b}) \sim P_{UV}$ \\
\STATE  Calculate the average and variance of the transformation predictions:
\STATE $m_{f} \leftarrow \frac{1}{b}\sum_{i=1}^{b}f_{\theta_f}(u_{i})$ ; $\sigma_{f}^{2} \leftarrow \frac{1}{b}\sum_{i=1}^{b}(f_{\theta_f}(u_{i})-m_{f})^2$
\STATE $m_{g} \leftarrow \frac{1}{b}\sum_{i=1}^{b}g_{\theta_g}(v_{i})$ ; $\sigma_{g}^{2} \leftarrow \frac{1}{b}\sum_{i=1}^{b}(g_{\theta_g}(v_{i})-m_{g})^2$
\STATE Standardize w.r.t. the minibatch: 
\STATE $\forall i:$ $\hat{f}_{\theta_f}(u_{i}) \leftarrow  \frac{f_{\theta_f}(u_{i})-m_{f}}{\sqrt{\sigma_{f}^{2}+\epsilon}} ; \hat{g}_{\theta_g}(v_{i}) \leftarrow  \frac{g_{\theta_g}(v_{i})-m_{g}}{\sqrt{\sigma_{g}^{2}+\epsilon}}$
\STATE Maximize the following objective function $J$ by gradient ascent:
\STATE $J(\theta_{f},\theta_{g})= \frac{1}{b}\sum_{i=1}^{b}\hat{f}_{\theta_f}(u_{i})*\hat{g}_{\theta_g}({v}_{i})$

\STATE $\theta_{f} \leftarrow \theta_{f} + \alpha_{f}\frac{\partial J(\theta_{f},\theta_{g})}{\partial \theta_{f}}$; \ $\theta_{g} \leftarrow \theta_{g} +\alpha_{g}\frac{\partial J(\theta_{f},\theta_{g})}{\partial \theta_{g}}$ 
\STATE \textbf{until} convergence
\end{algorithmic}
\label{alg:hgr_estimation}
\end{algorithm}

Algorithm \ref{alg:hgr_estimation} depicts the optimization process of \ref{opthgr}. Until convergence, it samples instantiations of $(U,V)$ from $P_{U,V}$ (or from a training set of data) to form mini-batches. At each iteration, it computes expectation and variance estimators of $f_{\theta_f}$ and $g_{\theta_g}$ on the current batch. These estimators are used to standardize the outputs of both neural networks on the batch. Finally, it updates the parameters of both networks by gradient ascent on the objective function to maximize $J(\theta_f,\theta_g)$. Note that the gradients are computed by back-propagating not only through the output values of $\theta_{f}$ and $\theta_{g}$ but also through means and variances of the batch, to ensure convergence. At the end, the $HGR_{\Theta}(U,V)$ estimator can be computed by considering the  expectation of the products of standardized outputs of both networks.

\begin{figure}[h]
  \centering
  \includegraphics[scale=0.25,valign=t]{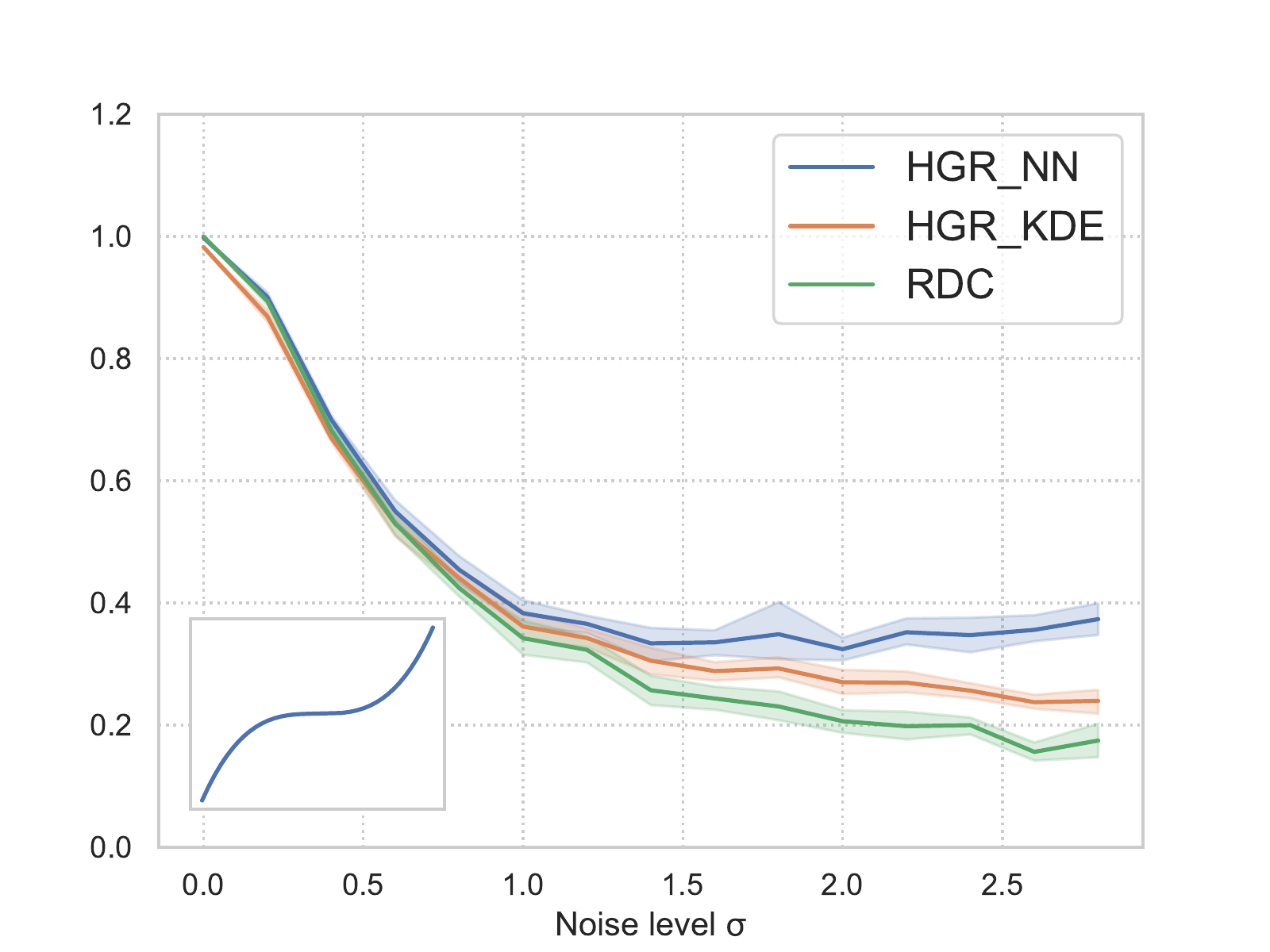}
  \includegraphics[scale=0.25,valign=t]{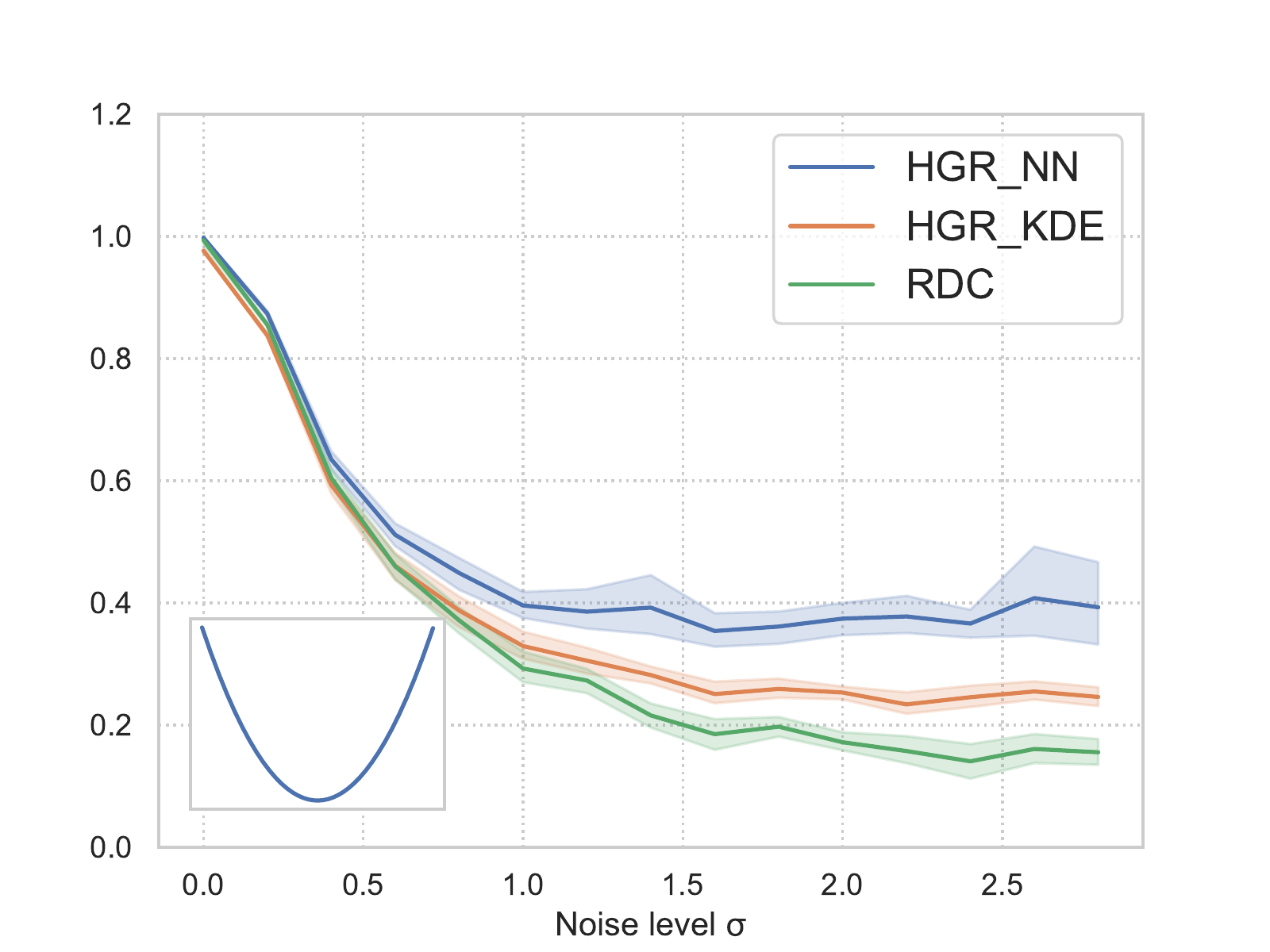}
  \includegraphics[scale=0.25,valign=t]{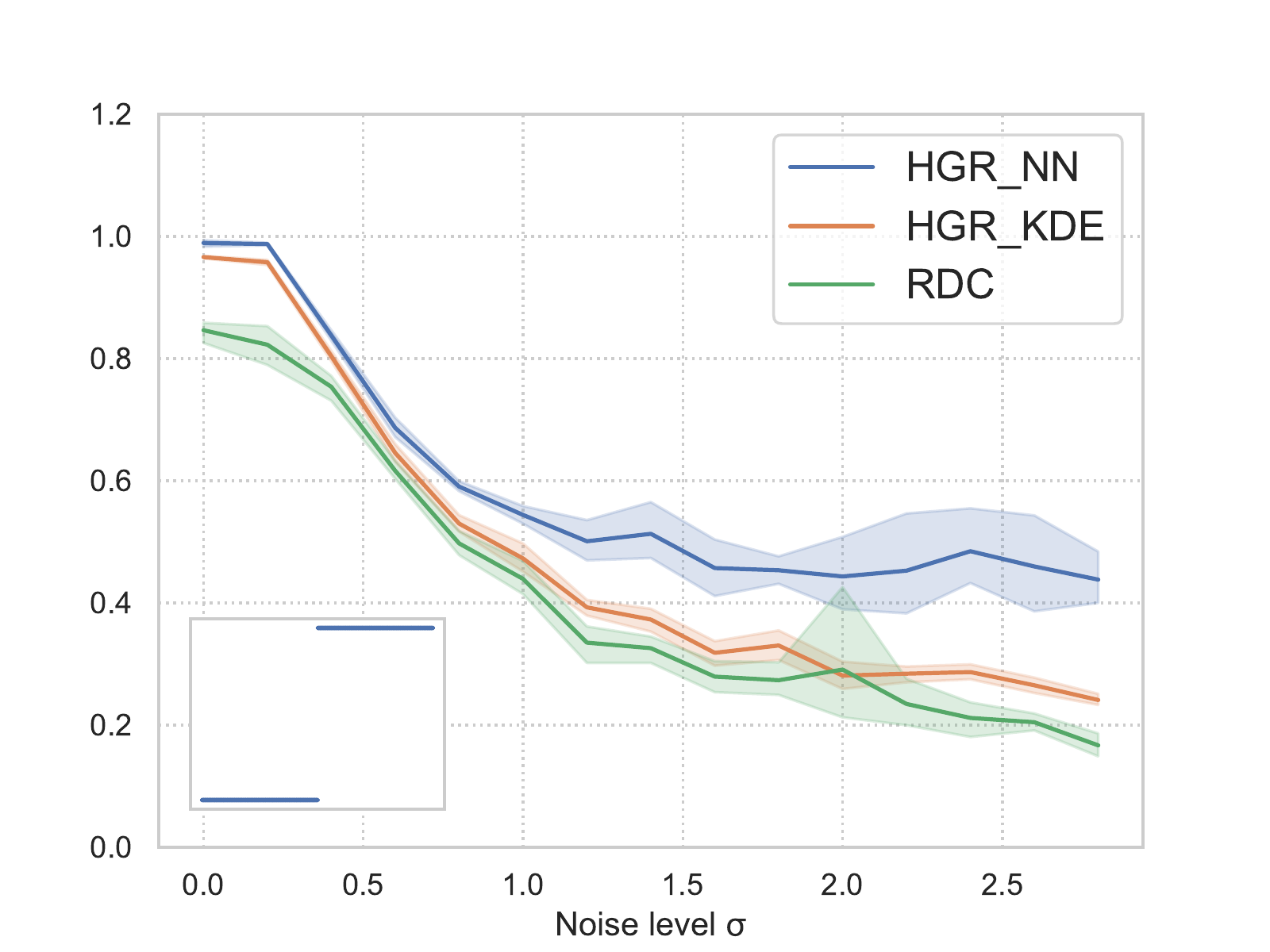}
  \includegraphics[scale=0.25,valign=t]{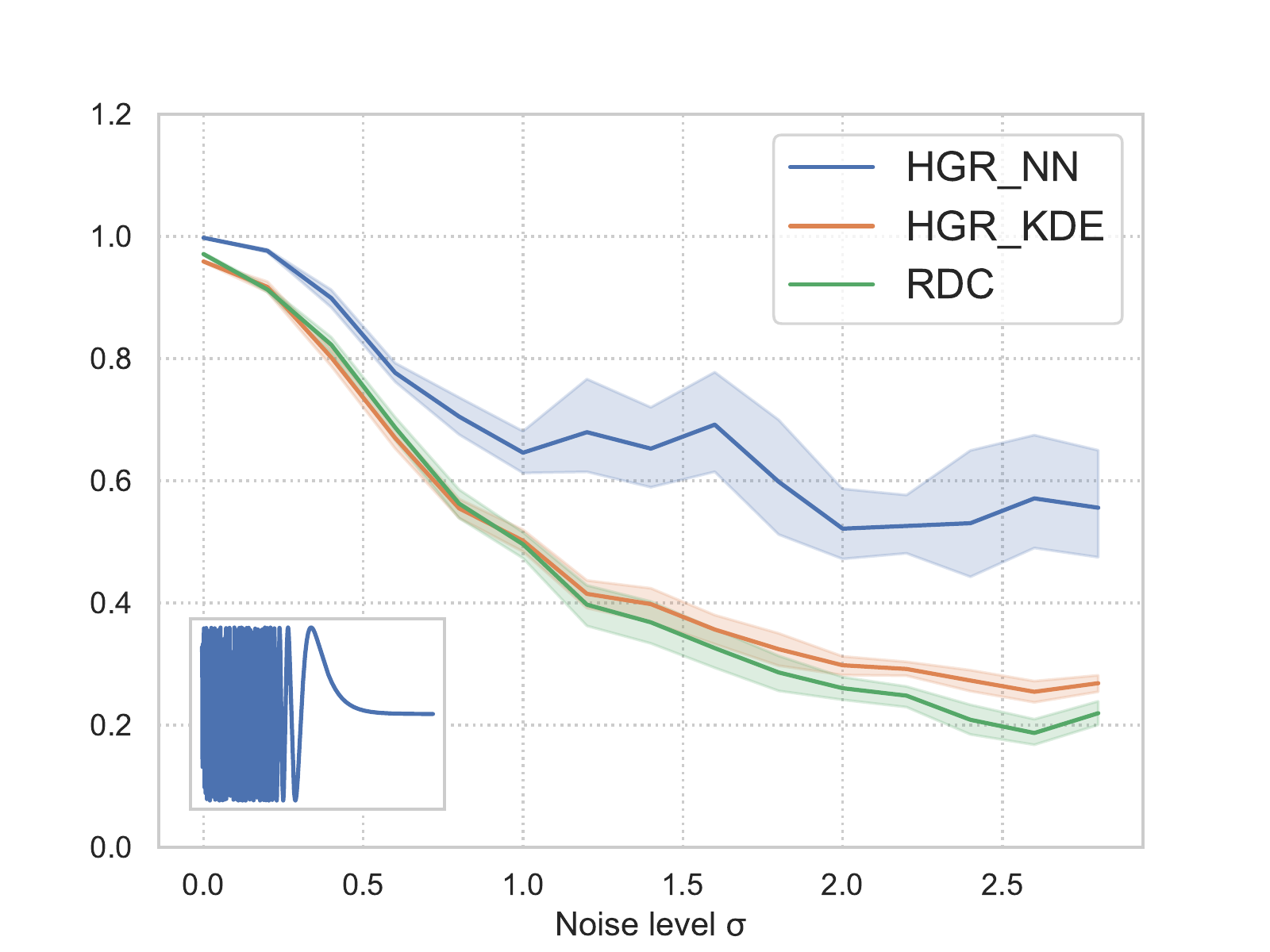}
  \caption{HGR estimation in various settings}
  \label{fig:empirical_assessments2}
\end{figure}

As $\chi^2_{\theta}$ was a lower-bound of $\chi^2$, the neural estimator $HGR_{\Theta}(U,V)$ is a lower-bound of $HGR(U,V)$ (at least on the training data set).  
However, as experimentally shown in figure \ref{fig:empirical_assessments2}, our estimator gives very accurate approximations in various settings. For these experiments, we produced artificial data $(U,V)$ with non-linear dependencies. Four data sets were generated by instantiating $U$ with samples drawn from a uniform distribution ${\cal U}(-10;10)$ and defining $V$ according to a non-linear transformation of $U$: $V=F(U)+\epsilon$, with $F$ a given association pattern and $\epsilon \sim {\cal N}(0,\sigma^2)$ a random noise added to $V$.  Each sub-figure in Fig.\ref{fig:empirical_assessments2} corresponds to a data set generated according to the association pattern plotted in the small box in its left corner (500 pairs $(U,V)$ were generated for each data set). 
Note that for each of the scenarios, the linear correlation between $U$ and $V$ is 0, but the HGR coefficient is theoretically equal to 1 (at least when no noise is added to the transformation). 
The aim is to assess the ability of the HGR estimators to recover the HGR value,  despite some complex association patterns and some noise in the data. We compare HGR\_KDE, HGR\_RDC and our estimator HGR\_NN, for which we consider neural networks $f$ and $g$ of three layers, each including ten units with $\tanh$ activation function and Xavier initialization\footnote{Note that $\chi^2$ measures cannot be plotted on the same figures, f-divergences being not comparable with correlations. When $HGR(U,V)=1$, $\chi^2(P_{U,V},P_{U} \otimes P_{V})=+\infty$.}. Results show that, when no noise is added to the data, HGR\_KDE and HGR\_RDC have difficulties to recover the optimal transformations on the two last scenarios in which the relationship is either not continuous or highly unsteady. 
Thanks to the higher freedom provided by the use of neural networks, HGR\_NN succeeds in retrieve a HGR of 1 for these settings. When noise is added to the data, the true HGR coefficient could  be lower than 1. We thus assess the ability of the estimators to approach the HGR value that would be induced by optimal transformations $f$ and $g$ on the data. 
Note that our approach cannot exceed its value, due to the use of a restricted set of neural transformation functions. From the figure, we observe that the curve of HGR\_NN is always the highest (thus the closest to the optimal value from the data), and that the difference between our approach and the others increases with noise. HGR\_NN appears more robust to noise. Additional experiments on the power of dependence of our estimator have also 
shown that our estimator is usually more efficient than its competitors for discerning dependent from independent samples in various settings. 

\begin{figure}[h]
  \centering
  \includegraphics[scale=0.75,valign=t]{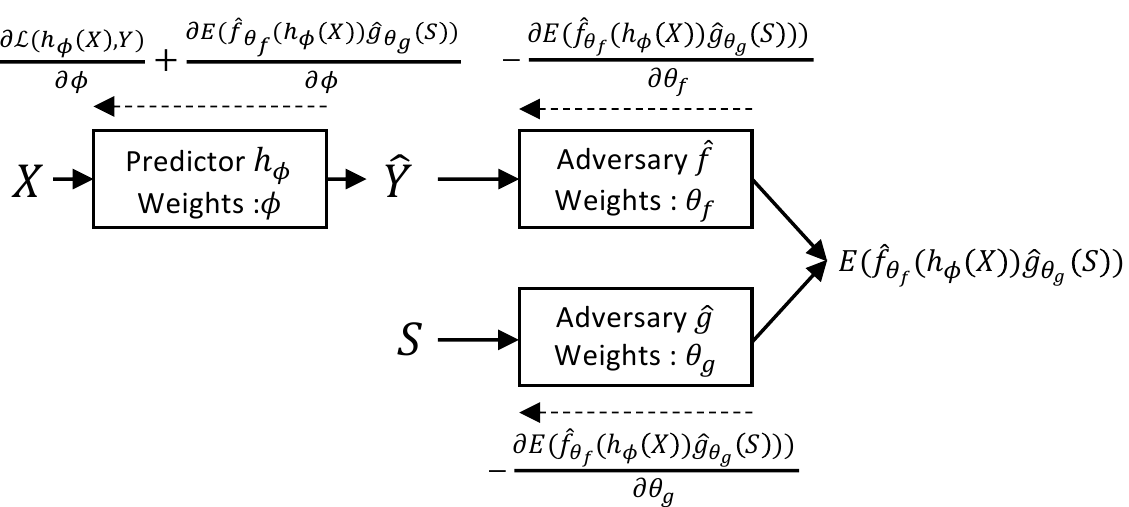}
  \caption{The Fair HGR NN adversarial algorithm for demographic parity.} 
  \label{fig:hgr_fair}
\end{figure}

Thus, our neural HGR estimator is a good candidate for standing as the adversary $\Psi(U,V)$ to plug in the global regression objective (\ref{genericfunction}). Figure \ref{fig:hgr_fair} gives the full architecture of our adversarial learning algorithm using the neural HGR estimator for demographic parity. It depicts the prediction function $h_{\phi}$, which outputs $\widehat{Y}$ from $X$, and the two neural networks $f_{\theta_f}$ and $g_{\theta_g}$, which seek at defining the more strongly correlated transformations of  $\widehat{Y}$ and $S$. Left arrows represent gradient back-propagation. As for our Fair $\chi^2$ NN algorithm, the learning of Fair HGR NN is done via stochastic gradient, alternating steps of adversarial maximization and global loss minimization.  The full Fair HGR NN algorithm is given in the supplementary material.

\section{Empirical Results}
\label{sec:results}


\subsection{Synthetic Scenario}

In order to test the efficiency of our algorithms, we set up a simple toy scenario. The subject is a pricing algorithm for a fictional household insurance policy. The goal of this exercise is to produce a fair predictor which estimates the average cost without incorporating any bias against the policyholder's age. We want to compare the two proposed algorithms (\textit{Fair HGR NN} and \textit{Fair $\chi^2$ NN}) with a classical neural network called \textit{Standard NN} and the algorithm \citep{mary2019fairness}. 


For the toy scenario, we create four explicit variables: Age of the policyholder, number of rooms, total surface, age of the building. We consider the policyholder's age as sensitive attribute and we construct the average cost variable $Y$ with the last three variables only (without the sensitive variable).  

To evaluate this, we create the target variable $Y$ with an exponential function which takes into account the three explicit variables mentioned above. The surface variable will be a trinomial transformation variable of age. This transformation is chosen such that no linear correlation with the surface variable exists (Pearson correlation = 0.00). On the other hand, given the non-linear relationship of the trinomial between the age and the surface variable, it is expected that the HGR coefficient will be non-zero for the Standard NN.




For each algorithm, we repeat five experiments by randomly
sampling two subsets, $80\%$ for the training set and $20\%$ for the test set. 



For the three mitigation algorithms, in order to solve this problem and, thus, to minimize the non-linear dependency between the age and the predictions we use a specific hyperparameter $\lambda$. This hyperparameter is obtained by 10-fold cross
validation on $20\%$ of the test set. The choice of this value depends on the main goal, resulting in a trade-off between accuracy and fairness. We decided to train a model which reaches a HGR coefficient of approximately $5\%$ for demographic parity and a HGR coefficient of $30\%$ for equalized residuals.

\begin{figure}[h]
  \centering
  \subfloat[Demographic Parity]{
    \includegraphics[scale=0.3,valign=t]{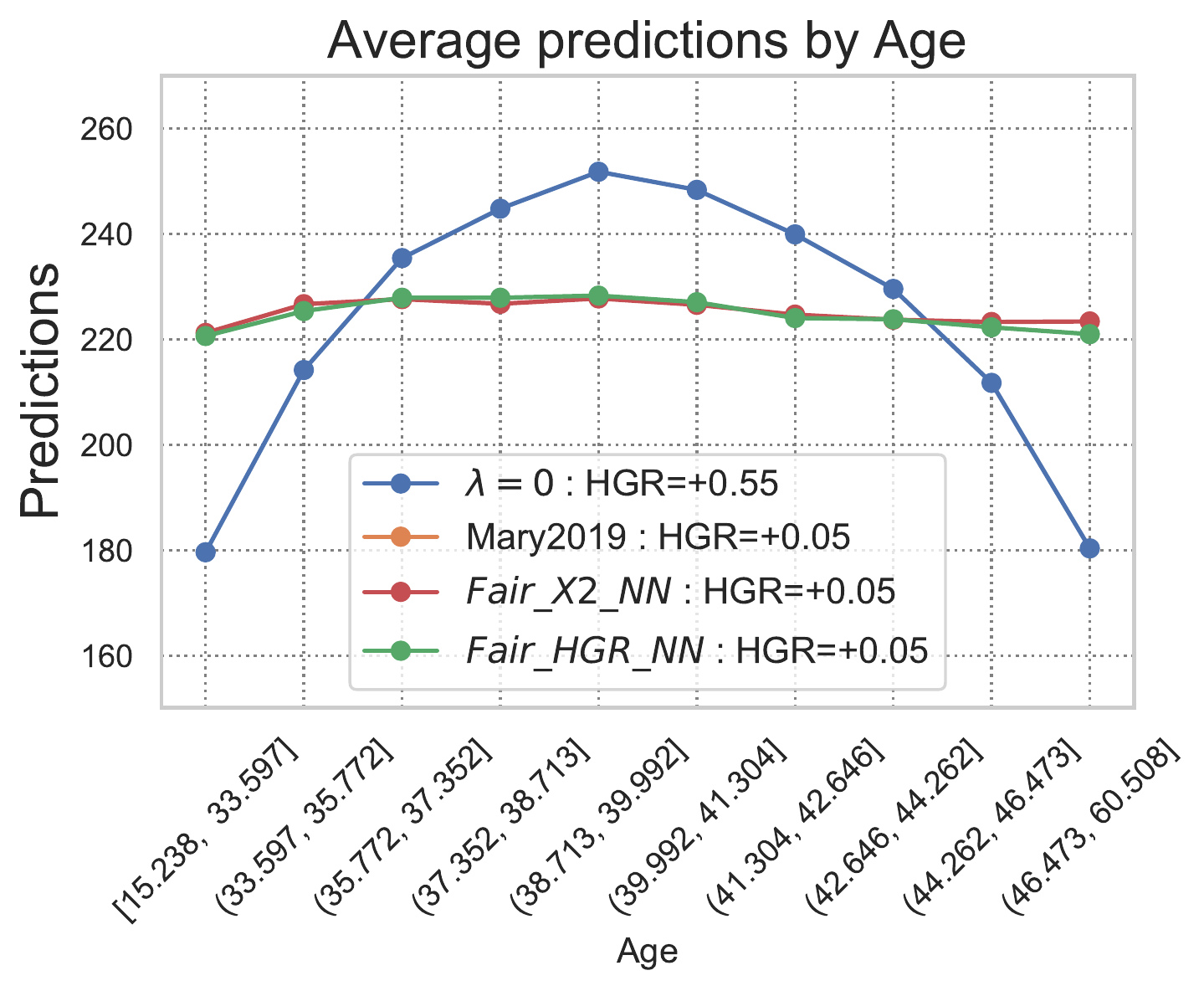}
  \label{fig:lambdapruleacc}
  }
    \subfloat[Equalized Residuals]{
    \includegraphics[scale=0.3,valign=t]{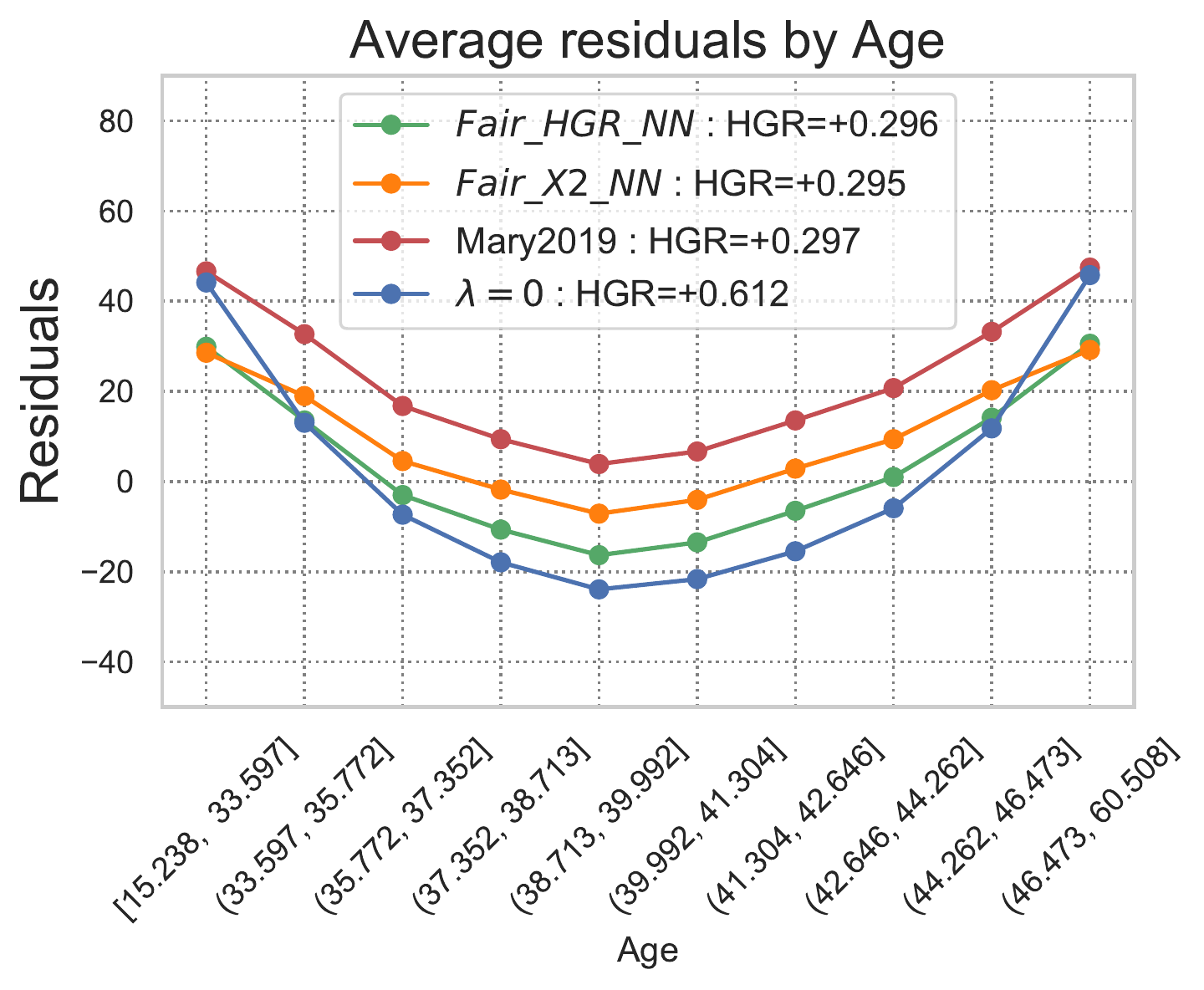}
  \label{fig:lambdapruleacc}
  }
  \caption{Synthetic scenario: Average predictions by age for demographic parity and average residuals for equalized residuals}
  \label{fig:synthetic_scenario}
\end{figure}

In Figure~\ref{fig:synthetic_scenario}, the blue curves represent the predictions of the Standard NN. The quadratic link between the prediction and the sensitive attribute age can be easily observed. 
In spite of this obvious (non-linear) link and the demographic disparity noticed, the linear Pearson coefficient is close to 0, which leads us to the necessity to minimize the HGR coefficient in order to increase the level of fairness. As expected, for demographic parity, the bias mitigation algorithms lead to price predictions almost stable around 226 euros. This attenuation is achieved with an increased MSE of about $10\%$. The results are quite similar for all three algorithms. 
For equalized residuals, the quadratic link between the residuals and the sensitive attribute age is also observed. The Standard NN obtains a high HGR of $0.61$. For a same level of fairness ($0.295$ of HGR), the Fair HGR NN increases the quadratic error by $67\%$, the Fair $\chi^2$ NN by $80\%$ and the \citep{mary2019fairness} algorithm by $98\%$.

\subsection{Real-world Experiments}

\begin{table*}[!h]
\caption{Results for Demographic Parity and Equalized Residuals in terms of accuracy (MSE) and fairness metrics. }
\label{tab:results_demographic_parity}
\centering

\resizebox{\textwidth}{!}{
\begin{tabular}{||l|l||l|l|l|l|l|l|l||l|l|l|l|l|l|l||}

\hhline{================} 
 \multicolumn{2}{||c||}{\multirow{2}{*}{ }} & \multicolumn{7}{c||}{Demographic Parity} & \multicolumn{7}{c||}{Equalized Residuals} \\ \cline{3-16}                                
  \multicolumn{2}{||c||}{ } & MSE & HGR\_NN & HGR\_KDE & HGR\_RDC & $\chi^2$\ KDE & $\chi^2$\ NN & FairQuant & MSE & HGR\_NN & HGR\_KDE & HGR\_RDC & $\chi^2$\ KDE & $\chi^2$\ NN & FairQuant \\ 
\hline
\multirow{5}{*}{\rotatebox[origin=c]{90}{US Census}}
& Standard NN & 0.274  & 0.212 & 0.181 & 0.217 & 0.036  & 0.032 &  0.059 &   0.274   &  0.157  &  0.098  &  0.122   & 0.012   &  0.010  & 0.008  \\
& Fair HGR NN& \textbf{0.526}  & \textbf{0.057} & \textbf{0.046} & \textbf{0.042} & 0.005 &   0.003 & \textbf{0.008} &  \textbf{0.334} & \textbf{0.068}   &  0.053  & \textbf{0.055} & \textbf{0.004}  &  \textbf{0.003}   &  \textbf{0.003} \\
& Fair $\chi^2$ NN & 0.535  & 0.069 & 0.048 & 0.044  &    \textbf{0.004} & \textbf{0.002} & \textbf{0.008}  &  0.384    &  0.084  &  0.054  & 0.057 & 0.005  &  \textbf{0.003} &   0.006  \\
& Mary2019~\cite{mary2019fairness}  & 0.541  & 0.075 & 0.061 & 0.078 & 0.005 & 0.003 & 0.019 &  0.408  &  0.092  & \textbf{0.049}  & 0.063  & 0.005 & \textbf{0.003}   &    0.009  \\ 
& Fair MINE NN & 0.537 & 0.058 & 0.048 & 0.045 & 0.005 & 0.003 & 0.012 &  0.406    &  0.083  &  0.055 & 0.057 & 0.005  &  \textbf{0.003} &  0.008  \\

\hline

\multirow{5}{*}{\rotatebox[origin=c]{90}{Motor}}
& Standard NN & 0.945   & 0.201 & 0.175 & 0.200 & 0.035 & 0.029 & 0.008   &  0.945     &  0.145 &  0.102  & 0.123 & 0.013  &  0.010 &   0.075  \\ 
& Fair HGR NN & \textbf{0.971}   & \textbf{0.072} & \textbf{0.058} &  \textbf{0.066} & \textbf{0.006} & \textbf{0.004} & \textbf{0.002}  &   \textbf{0.991}  & \textbf{0.102}  &  0.082   & 0.092 & 0.010  &   0.008  &  \textbf{0.011} \\ 
& Fair $\chi^2$ NN & 0.975  & 0.076 & 0.067 & 0.067 & 0.008 & 0.005 & 0.004 &    1.011  & 0114   &   0.081 & \textbf{0.077} &  0.010  &  0.008 &   0.014 \\ 
& Mary2019~\cite{mary2019fairness}  & 0.979   & 0.077 & 0.059 & 0.067 & \textbf{0.006} & 0.005 & 0.004  &   1.019    &   0.111 &  \textbf{0.079}  & 0.098  &  \textbf{0.009}   &    \textbf{0.006}  & 0.015  \\ 
& Fair MINE NN & 0.982 & 0.078 & 0.068 & 0.069 & 0.008 & 0.006 & 0.004 &    1.024  & 0.121   &   0.091 & 0.092 &  0.012  &  0.012 &  0.031 \\

\hline

\multirow{5}{*}{\rotatebox[origin=c]{90}{Crime}}
& Standard NN                        &    0.384     &  0.732 &  0.525 &  0.731 & 0.319  &    0.352 &  0.353   &   0.385  &  0.472   &  0.244    & 0.440 & 0.092  &  0.096 & 0.047  \\ 
& Fair HGR NN    &     0.781     &  \textbf{0.356} &  \textbf{0.097}  &  \textbf{0.171}  &  \textbf{0.014}   & \textbf{0.064} & \textbf{0.039}  &   0.583    & 0.382   &    0.151 & 0.222 & 0.038  & 0.092   & \textbf{0.028}          \\ 
& Fair $\chi^2$ NN  &   0.785     &  0.385 &  0.106  &  0.184  &  0.020   & 0.079 & 0.123  &   0.581    & \textbf{0.353}   &    \textbf{0.142} & \textbf{0.211} & \textbf{0.035}  & \textbf{0.090}   & 0.042     \\ 
& Mary2019~\cite{mary2019fairness}    &   \textbf{0.778}  &   0.371    &  0.115 &  0.177  &  0.022   &   0.090 &      0.064  &   \textbf{0.579}    & 0.381   &    0.152 & 0.221 & 0.038  & 0.092   & 0.048    \\ 
& Fair MINE NN  &   0.782     &  0.395 &  0.110  &  0.201 &  0.022   & 0.079 & 0.136 &   0.583    & 0.413  &    0.161 & 0.232 & 0.041  & 0.098  & 0.038 \\

\hhline{================} 
\end{tabular}
}
\centering
\\
\end{table*}

Our experiments on real-world data are performed on the three following data sets: 
\begin{itemize}
\item The US Census demographic data set is an extraction of the 2015 American Community Survey 5-year estimates. It contains 37  information features about 74,000 American census tracts. Our goal is to predict the percentage of children below the poverty line. We consider gender as a sensitive attribute encoded as the percentage of the women in the census tract.
\item The Motor Insurance data set originates from a pricing game organized by The French Institute of Actuaries in 2015~\citep{pricinggame15}. 
The data set contains a total of 15 attributes for 36,311 observations. The task is to predict the average claim cost of third-party material claims per policy. As the sensitive attribute we use the driver's age. 
\item The Crime data set is obtained from the UCI Machine Learning Repository~\citep{Dua:2019}. This data set includes a total of 128 attributes for 1,994 instances from communities in the US. The task is to predict the number of violent crimes per population for US communities. As the sensitive attribute we use the race information with the ratio of an ethnic group per population.
\end{itemize}
For all data sets, we repeat five experiments by randomly sampling two subsets, 80\% for the training set and 20\% for the test set. Finally, we report the average of the mean squared error (MSE), and the mean of the fairness metrics HGR\_NN, HGR\_KDE, HGR\_RDC, $\chi^2$KDE, $\chi^2$NN from the test set. Since none of these fairness measures are fully reliable (they are only estimations which are used by the compared models), we also introduce a metric based on discretization of the sensitive attribute. This $FairQuant$ metric splits the test samples in 50 quantiles with regards to the sensitive attribute, in order to obtain sample groups of the same size. For each of them, we compute the mean of $h(X)$ for demographic parity, of $h(X)-Y$. for equalized residuals. Finally, $FairQuant$ equals the mean absolute difference between the global average and the means computed in each quantile (e.g., for demographic parity, $FairQuant=\frac{1}{50}\sum_{i=1}^{50} |m_i - m|$, with $m_i$ the mean of $h(X)$ in the $i$-th quantile and $m$ its mean on the full test set).          
As a baseline, we use a classic, "unfair" deep neural network, Standard NN. We compare our two approaches with a similar approach that would use mutual information rather than HGR in our framework (see section \ref{sec:HGR_Adversarial_Network})  with Mary2019~\citep{mary2019fairness}~\footnotemark[2], which considers the minimization of $\chi^2$\_KDE as described above. 

\footnotetext[2]{\url{https://github.com/criteo-research/continuous-fairness}}


For each algorithm and for each data set, we obtain the best hyperparameters by grid search in five-fold cross validation (specific to each of them). Depending on the task, we parameterized the number of layers between 3 and 5 and between 8 and 32 for the number of units. We used Tanh activation functions,  Dropout and Xavier initialization. The considered regression loss is MSE.


Results of our experiments can be found in Table~\ref{tab:results_demographic_parity}.  
For all of them, we attempted to obtain comparable results by giving similar accuracy of the models (MSE) in a same setting, via the hyperparameter $\lambda$ of our models that allows us to balance the relative importance of accuracy and fairness while learning. 

As expected, the baseline, Standard NN, is as expected the best predictor but also the most biased one. It achieves the lowest prediction errors and ranks amongst the highest and thus worst values for all fairness measures throughout all data sets and tasks.

For demographic parity, Fair HGR NN achieves the best accuracy and the best level of fairness assessed by HGR estimation and FairQuant. 
The Fair $\chi^2$ NN performs slightly worse, while the decrease in $\chi^2$ is not as much reflected in results for HGR. For example, on the US Census data the estimated $\chi^2$ is smaller than the Fair HGR NN but it still achieves worse results on the FairQuant metric (which is the most reliable fairness metric, since others are only estimations). Except on the Crime data set, Fair $\chi^2$ NN achieves better results for accuracy and fairness than Mary2019. The small amount of data seems to make it difficult to estimate $\chi^2$ (only 2,000 observations for Crime). It’s for this data set that we observe a significant deviation of results between $\chi^2$ KDE and $\chi^2$ NN. This may explain the worse results in terms of $\chi^2$ NN.


For equalized residuals, Fair HGR NN achieves the lowest values for the metric FairQuant for all three data sets (like for demographic parity). The Fair $\chi^2$ NN performs slightly worse. For MINE, except for UC Census, it achieves worse results in fairness and accuracy. Just like for demographic parity, it seems difficult to estimate KDE because the Crime data set is so small (this is highlighted by the significant difference between average results in terms of $\chi^2$ KDE and KDE NN on this data set). Globally, our neural approaches, and especially Fair HGR NN, appear very competitive in every setting.  

\label{sec:experiments}
\section{Conclusion}
\label{sec:conclusion}
We developed two new adversarial learning approaches to produce fair continuous predictions with a continuous sensitive attribute. 
The first method shows an improvement of the approach presented in \cite{mary2019fairness} with an estimation and minimization of the $\chi^2$ upper bound directly by neural network. In the second approach we propose 
to reduce the HGR coefficient by estimating it via standardized neural networks. This method proved to be very efficient for two fairness objectives on various artificial and real-world data sets. 



\clearpage

\bibliography{references}
\bibliographystyle{aaai}

\end{document}


%
\title{Fairness-Aware Neural R\'eyni Minimization for Continuous Features \\ {\normalsize Supplementary Material}}

\maketitle
\begin{abstract}
This is the supplementary material of paper Fairness-Aware Neural R\'eyni Minimization for Continuous Features. 
\end{abstract}

\section{Fair algorithms}

This section gives details for the fair algorithms we propose in the paper. 

Algorithm \ref{alg:fair_chi2} depicts our Fair $\chi^2$ NN algorithm for the Demographic Parity task (for the Equalized Residuals task, the only change is in the function $f$ that takes $h_{\phi}(x_{i}) - y_i$ as input rather than simply $h_{\phi}(x_{i})$). The algorithm takes as input a training set from which it samples batches of size $b$ at each iteration. It also draws from the training set $b$ samples of the sensitive $\bar{s}_1, . . . ,\bar{s}_b$ to obtain  samples of $S$ independent from $X$. Then it computes the objective function to maximize to estimate the $\chi^2$-divergence and the global regression objective. Finally, at the end of each iteration, the algorithm updates the parameters of the adversary $\theta$ by one step of gradient ascent and the regression parameters $\phi$ by one step of gradient descent. 

\begin{algorithm}
\caption{Fair $\chi^2$ NN for Demographic Parity}
\begin{algorithmic} 

\STATE \textbf{Input:}
Training set ${\cal T}$, 
Loss function $\mathcal{L}$, Batchsize $b$, $\quad$
{\color{white} \textbf{Input:    }} Neural Networks $h_{\phi}$ and $f_{\theta}$,  $\qquad$ $\qquad$ $\quad$ 
{\color{white} \textbf{Input:   }}  Learning rates $\alpha_f$ and $\alpha_h$, Fairness control $\lambda$ 
\STATE \textbf{Repeat}
\STATE Draw $b$ samples $(x_{1}, s_1, y_{1}), . . . ,(x_{b}, s_b, y_{b})$ from ${\cal T}$
\STATE Draw $b$ samples $\bar{s}_1, . . . ,\bar{s}_b$ from ${\cal T}$
\STATE Compute the objectives:
\STATE $J(\theta) \leftarrow \frac{1}{b} \sum_{i=1}^{b} f_{\theta}(h_{\phi}(x_{i}),s_{i})-$\\
$\qquad \quad \  \frac{1}{b}\sum_{i=1}^{b}(f_{\theta}(h_{\phi}(x{i}),\bar{s}_{i})+\frac{1}{4}f_{\theta}(h_{\phi}(x_{i}),\bar{s}_{i})^2)$
\STATE $L(\phi,\theta) = \frac{1}{b}\sum_{i=1}^{b} \mathcal{L}(h_{\phi}(x_i),y_i) + \lambda J(\theta)$
\STATE Update the adversary by gradient ascent: 
\STATE $\theta \leftarrow \theta + \alpha_f \frac{\partial J(\theta)}{\partial \theta}$ 
\STATE Update the predictor model $h_{\phi}$ by gradient descent:
\STATE $\phi \leftarrow \phi -\alpha_{h} (\frac{\partial L(\phi,\theta)}{\partial {\phi}})$

\end{algorithmic}
\label{alg:fair_chi2}
\end{algorithm}

Algorithm \ref{alg:fair_hgr} depicts our Fair HGR NN algorithm for the Demographic Parity task (as for $\chi^2$, the only change for Equalized Residuals is in the input of function $f$). The algorithm takes as input a training set from which it samples batches of size $b$ at each iteration. At each iteration it first standardize the output scores of networks $f_{\theta_f}$ and $g_{\theta_g}$ to ensure 0 mean and a variance of 1 on the batch. Then it computes the objective function to maximize to estimate the HGR score and the global regression objective. Finally, at the end of each iteration, the algorithm updates the parameters of the adversary $\theta_f$ and $\theta_g$ by one step of gradient ascent and the regression parameters $\phi$ by one step of gradient descent. Back-propagation is performed on the full architecture, including means and variance calculations, to avoid oscillations.

\begin{algorithm}
\caption{Fair HGR NN for Demographic Parity}
\begin{algorithmic} 

\STATE \textbf{Input:}
Training set ${\cal T}$, 
Loss function $\mathcal{L}$, Batchsize $b$, $\quad$
{\color{white} \textbf{Input:    }} Neural Networks $h_{\phi}$, $f_{\theta_f}$ and $g_{\theta_g}$,  $\qquad$ $\qquad$ $\quad$ 
{\color{white} \textbf{Input:   }}  Learning rates $\alpha_f$, $\alpha_g$ and $\alpha_h$, Fairness control $\lambda$ 
\STATE \textbf{Repeat}
\STATE Draw $b$ samples $(x_{1}, s_1, y_{1}), . . . ,(x_{b}, s_b, y_{b})$ from ${\cal T}$
\STATE Calculate the mean and variance of the transformations:  
\STATE $m_{f} \leftarrow \frac{1}{b}\sum_{i=1}^{b}f_{\theta_f}(h_{\phi}(x_i))$ ; $m_{g} \leftarrow \frac{1}{b}\sum_{i=1}^{b}g_{\theta_g}(s_{i})$  
\STATE $\sigma_{f}^{2} \leftarrow \frac{1}{b}\sum_{i=1}^{b}(f_{\theta_f}(h_{\phi}(x_i))-m_{f})^2$
\STATE $\sigma_{g}^{2} \leftarrow \frac{1}{b}\sum_{i=1}^{b}(g_{\theta_g}(s_{i})-m_{g})^2$ \\
\STATE Standardize the transformations:
\STATE $\forall i: \hat{f}_{\theta_f}(h_{\phi}(x_i)) \leftarrow  \frac{f_{\theta_f}(h_{\phi}(x_i))-m_{f}}{\sqrt{\sigma_{f}^{2}+\epsilon}} $
\STATE $\forall i: \hat{g}_{\theta_g}(s_i) \leftarrow  \frac{g_{\theta_g}(s_i)-m_{g}}{\sqrt{\sigma_{g}^{2}+\epsilon}}$
\STATE Compute the objectives:
\STATE $J(\theta_{f},\theta_{g})=\frac{1}{b}\sum_{i=1}^{b}\hat{f}_{\theta_f}(h_{\phi}(x_i))*\hat{g}_{\theta_g}({s}_{i})$
\STATE $L(\phi,\theta_{f},\theta_{g}) = \frac{1}{b}\sum_{i=1}^{b} \mathcal{L}(h_{\phi}(x_i),y_i) + \lambda J(\theta_{f},\theta_{g})$
\STATE Update the adversary by gradient ascent: 
\STATE $\theta_{f} \leftarrow \theta_{f} + \alpha_{f}\frac{\partial J(\theta_{f},\theta_{g})}{\partial \theta_{f}}$; \ $\theta_{g} \leftarrow \theta_{g} +\alpha_{g}\frac{\partial J(\theta_{f},\theta_{g})}{\partial \theta_{g}}$ 
\STATE Update the predictor model $h_{\phi}$ by gradient descent:
\STATE $\phi \leftarrow \phi -\alpha_{h} (\frac{\partial L(\phi,\theta_{f},\theta_{g})}{\partial {\phi}})$

\end{algorithmic}
\label{alg:fair_hgr}
\end{algorithm}

\section{Mutual Information vs $\chi^2$}

We give here a short analysis of dominance relationships between HGR, $\chi^2$ and the mutual information in the bivariate gaussian case.  

\begin{proposition}
By considering $(U,V)$ be jointly Gaussian variables, we have:
$HGR(U,V)^2 \leq 1-2^{-2*I(U,V)}$, and because the mutual information can always be upper bounded as: 
$I(U,V) \leq log(1+\chi^{2})$, we have:
\begin{eqnarray}
    \left\{
    \begin{array}{ll}
        HGR(U,V)^2 \leq 1-2^{-2*I(U,V)} \leq \chi^{2} ; \forall \chi^{2} \geq t \\
        HGR(U,V)^2 \leq \chi^{2} \leq 1-2^{-2*I(U,V)} ;  \mbox{otherwise}
    \end{array}
\right.
\end{eqnarray}
with $t=\frac{e^{\frac{-1}{2log(2)}}-1}{1+e^{\frac{-1}{2log(2)}}} \approx 0.345$ and $\chi^2=\chi^2(P_{UV},P_{U} \otimes P_{V})$
\end{proposition}
\begin{proof}
The chi-square divergence is an upper bound of the Kullback-Leibler divergence:
$\infdiv{P_{U}}{P_{V}} \leq \chi^2(P_{U},P_{V})$
with:
\begin{eqnarray}
\infdiv{P_{U}}{P_{V}} =  \int_{\mathbb R}\int_{\mathbb R} P_{U}(j,j^{'})*log(\frac{P_{U}(j,j^{'})}{Q_{U}(j,j^{'})}) djdj'
\end{eqnarray}
The proof of this inequality follows straightforwardly from
Jensen’s inequality and the concavity of the logarithm.
Because $I(U,V)= \infdiv{P_{U}}{P_{V}}$, the mutual information can always be upper bounded as:

\begin{eqnarray}
I(U,V) \leq \chi^2(P_{UV},P_{U} \otimes P_{V})
\end{eqnarray}

In the gaussian case, it has been proven that the mutual information is an upper bound of the HGR \citep{asoodeh2015maximal}: $HGR(U,V)^2 \leq 1-2^{-2*I(U,V)}$.
Then the calculation of the threshold $t$ is straightforward.


\end{proof}

\section{Synthetic scenario}
We report below details on the distributions used in the synthetic scenario:
\begin{eqnarray*}
\begin{aligned}
Age \sim& \mathcal{N}(40,5)\\
Rooms \sim& \left\lfloor\mathcal{U}(1,5)\right\rfloor\\
\epsilon \sim& \mathcal{N}(0,1)\\
Surface =&-0.25*(-Age+40)^{2}+120+\epsilon\\
BldgAge \sim& \mathcal{N}(30,10)\\
Y =& 0.0005*e^{(0.07*Surface+0.08*BldgAge+0.4*MRooms)}\\
  & +150
\end{aligned}
\end{eqnarray*}

\bibliography{references}
\bibliographystyle{aaai}